\documentclass[11pt]{article}
\PassOptionsToPackage{breaklinks,hidelinks}{hyperref}

\usepackage[preprint]{acl}

\usepackage{times}
\usepackage{latexsym}

\usepackage[T1]{fontenc}

\usepackage[utf8]{inputenc}

\usepackage{microtype}

\usepackage{inconsolata}

\usepackage{graphicx}

\usepackage[inline]{enumitem} 
\setlist[itemize]{leftmargin=*, nosep}
\setlist[enumerate]{leftmargin=*,nosep,label=(\arabic*)}

\usepackage{booktabs}
\usepackage{multirow}
\usepackage{tabularx}
\usepackage{makecell}
\usepackage{array}

\usepackage{xcolor}
\usepackage{tikz}
\usetikzlibrary{calc,arrows.meta,fit}
\usepackage[hidelinks]{hyperref}

\usepackage{colortbl}
\usepackage[most]{tcolorbox}

\definecolor{accentblue}{HTML}{2C5F8A}
\definecolor{rowgray}{HTML}{F4F7FB}
\definecolor{mydarkblue}{RGB}{13,54,117}
\definecolor{myboxbg}{RGB}{248,251,255}

\newtcolorbox[auto counter, number within=section]{workedexamplebox}[2][]{
  enhanced, breakable,
  colback=myboxbg,
  colframe=mydarkblue,
  boxrule=1pt,
  arc=1mm,
  boxsep=0.6mm,
  left=0.8mm, right=0.8mm, top=0.6mm, bottom=0.4mm,
  before skip=4pt, after skip=4pt,
  title={Box~\thetcbcounter: #2},
  fonttitle=\bfseries\footnotesize,
  fontupper=\footnotesize,
  label={box:worked-example}
}

\usepackage{float}

\DeclareRobustCommand{\revise}[1]{\textcolor{black}{#1}}
\DeclareRobustCommand{\rev}[1]{\textcolor{black}{#1}}
\usepackage{acronym}
\acrodef{AI}{artificial intelligence}
\acrodef{LLM}{large language model}
\acrodef{SOTA}{state-of-the-art}
\acrodef{HCI}{human computer interaction}
\acrodef{MHS}{mental health support}

\newcommand{\OurModel}[1]{WikiHowAgent}

\title{\revise{Calibrating Trust in Mental Health AI:\\ A Three-Layer Survey across Stakeholders}}

\title{Surveying the Landscape Toward Trustworthy Mental Health AI: Calibrating Trust across Human, AI, and Interaction Layers}

\title{Trust Stack for Mental Health AI: A Survey of Calibration\\across Human, Interaction, and AI Layers}

\author{
     \textbf{Xin Sun\textsuperscript{1,2}},
     \textbf{Yue Su\textsuperscript{3}},
     \textbf{Yifan Mo\textsuperscript{3}},
     \textbf{Qingyu Meng\textsuperscript{3}},
     \textbf{Yuxuan Li\textsuperscript{3}},
     \textbf{Min Chen\textsuperscript{3}},\\
     \textbf{Mengyuan Zhang\textsuperscript{3}},
     \textbf{Saku Sugawara\textsuperscript{1}},
     \textbf{Charlotte Gerritsen\textsuperscript{3}},\\
     \textbf{Sander L. Koole\textsuperscript{3}},
     \textbf{Koen Hindriks\textsuperscript{3}},
     \textbf{Jiahuan Pei\textsuperscript{3}}
     \medskip
     \\ 
     \textsuperscript{1}National Institute of Informatics (NII), Japan\\
     \textsuperscript{2}University of Amsterdam, the Netherlands\\
     \textsuperscript{3}Vrije Universiteit Amsterdam, the Netherlands\\
}

\begin{document}
\maketitle
\begin{abstract}
Language-based AI is increasingly deployed for mental health support, yet trust is evaluated in interdisciplinary but operationally misaligned ways: NLP and AI work measures robustness, safety, privacy, and explanations, while psychotherapy, HCI, and regulatory work emphasize therapeutic fidelity, lived experience, empathy, and reliance. Empathetic chatbots can elicit strong user trust without commensurate safety, while safer systems are under-trusted when their boundaries are opaque, a calibration gap no single community owns. 
Through a structured scoping synthesis of 61 papers, we survey this landscape into a three-layer framework separating (L1) human-oriented trust, (L2) interaction-oriented trustworthiness, and (L3) AI-oriented trustworthiness, and map five stakeholder perspectives onto these layers. 
We outline a research agenda for building socio-technically aligned trustworthy AI for mental health support, highlighting that the central objective should shift from maximizing perceived trust to calibrating human trust to demonstrated interaction- and AI-level trustworthiness.

\end{abstract}


\section{Introduction}




\revise{Building trustworthy \ac{AI} systems for mental support is an interdisciplinary problem. 
Recent advances in \acfp{LLM} have expanded applications ranging from screening and psychoeducation to conversational support~\cite{na-etal-2025-survey}.
Recent work spans clinical, natural language processing (NLP), \rev{\ac{HCI}}, safety, and regulatory communities (see Appendix~\ref{appendix:methodology}, \autoref{fig:example} for a discipline-network view of 1,706-paper mapping corpus). 
Yet this literature is not isolated; rather, it is operationally misaligned: different communities often use similar trust language while evaluating different objects, such as users' subjective trust, interactive behavior, or system-level trustworthiness.}

\revise{This misalignment matters because trust and trustworthiness are related but not equivalent. Trust is a trustor's subjective attitude and willingness to rely on a system, whereas trustworthiness refers to evaluable properties of the trustee, such as reliability, safety, privacy protection, fairness, and accountability~\cite{jacovi2021formalizing,lee2004trust}. 
In mental health, the distinction is consequential: empathetic or human-like interaction can increase perceived trust even when AI model behavior is unstable, while strong safeguards may still fail if users cannot understand system limits. 
}

Prior surveys provide important foundations but leave three gaps that we address. 
First, general human-AI trust surveys~\cite{mehrotra2024systematic,regona2026building} synthesize trust theories across domains but do not engage with mental-health-specific pressures such as clinical confidentiality, crisis escalation, vulnerability, or over-disclosure. 
Second, medical-AI surveys~\cite{zhu-etal-2025-trust} focus on hallucination but treat trust as a property of model accuracy rather than a multi-stakeholder problem. 
Third, psychotherapy-focused LLM surveys~\cite{na-etal-2025-survey} catalog applications and risks but do not separate subjective trust, interaction- and system-level trustworthiness, making it difficult to diagnose where trust failures originate. 
Our framework addresses the calibration gap by mapping evidence to three layers, allowing trust failures to be attributed to specific stakeholder responsibilities. 
\autoref{tab:mental-health-specificity} further details how mental-health pressures reshape general human-AI trust evaluation.

\rev{Specifically, we ask: how should the field evaluate whether users' trust in language-based mental health AI is appropriate?}
We address this gap with a three-layer trust framework for language-based \ac{MHS} systems. The framework distinguishes human-oriented trust (users' subjective trust, expectations, and reliance), interaction-oriented trustworthiness (user-visible behaviors that communicate capability, boundaries, uncertainty, empathy, and control), and AI-oriented trustworthiness (model- and system-level properties that justify trust). We map these layers to psychotherapy, HCI, AI/NLP, safety/security, and regulatory perspectives, while treating clients and end users as the primary trustors represented in the human-oriented layer. 
Details of the PRISMA-guided review~\rev{\citep{PRISMA}} (not a database-exhaustive systematic review) and coding procedure are provided in \autoref{appendix:methodology}\rev{, including the PRISMA flow diagram (separate stratified search of 735 records; 61-paper analytic corpus)}.


The contributions of this survey are threefold.
\begin{itemize}[leftmargin=*, itemsep=0pt]
    \item \revise{We formalize trust calibration in mental health AI by distinguishing trust, trustworthiness, and reliance across three evaluation layers.}

    \item \revise{We conduct a PRISMA-guided scoping review of recent language-based mental health AI literature through this framework, clarifying how papers are searched, screened, coded, and mapped to stakeholder evaluation paradigms.}

    \item We derive an agenda to examine current methodology and evaluation across layers and stakeholders, positioning the ``gap'' that links AI advances to interaction behavior and human trust.
\end{itemize}

\vspace{-0.0mm}





\section{Conceptual Framework}


\revise{Trustworthy MHS AI is not a monolithic concept. It is multi-stakeholder evaluation problem in which human trust, interaction behavior, model capability, safety governance and responsibility must be examined together.}
\revise{We distill core trustworthiness \textit{criteria} (e.g., \rev{robustness}, faithfulness, transparency) into a three-layer trust framework, and review this landscape of involved \textit{stakeholders} (i.e., practitioners, AI/HCI/safety researchers, regulators).}

\begin{figure}[!ht]
    \centering
    \includegraphics[width=0.84\linewidth]{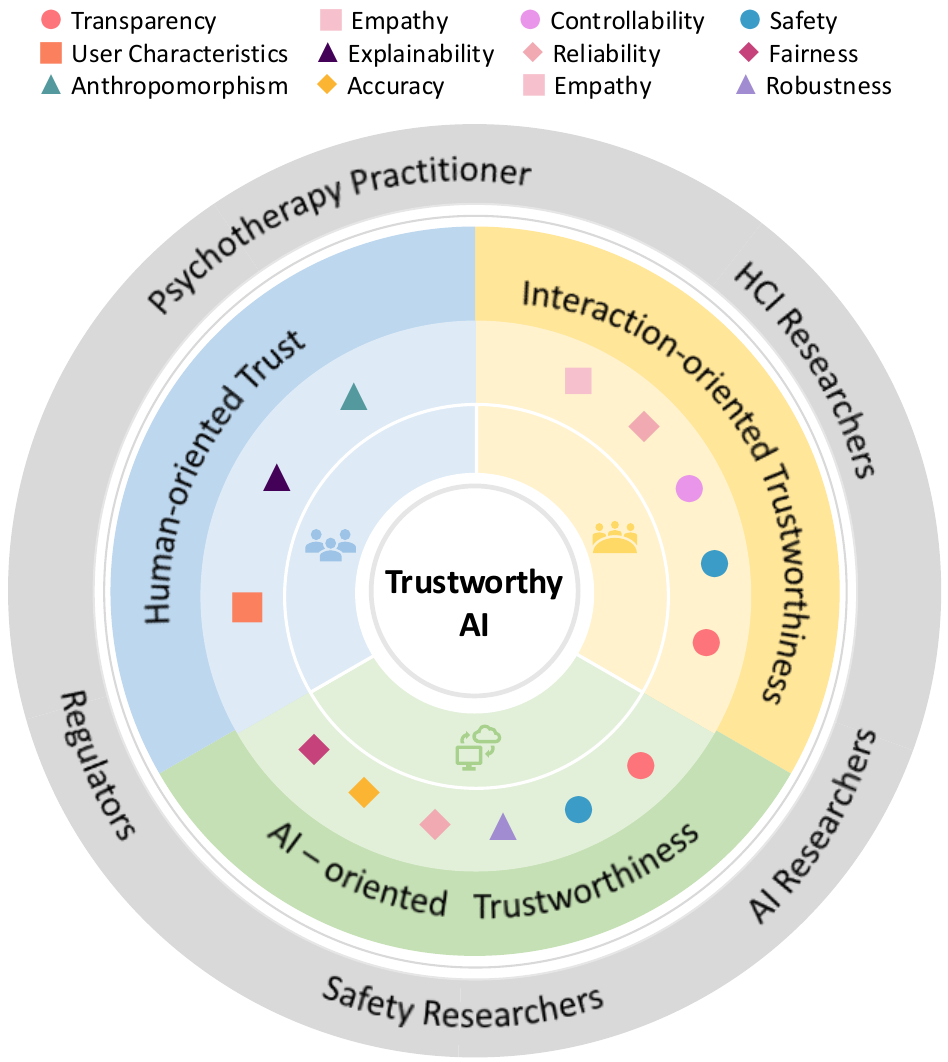}
    \vspace{-0.20mm}
    \caption{Based on a cross-disciplinary literature review, we identify five research and governance perspectives that shape trustworthy mental health AI. 
    We synthesize these perspectives into three layers: human-oriented trust, interaction- and AI-oriented trustworthiness.
    Clients, patients, and end users are the primary trustors represented in the human-oriented layer and defined in \autoref{tab:stakeholder-mapping}; the outer sectors indicate evaluation perspectives, not exclusive layer ownership. 
    }
    \label{framework}
    \vspace{-0.20mm}
\end{figure}


\subsection{Definitions and Background}
\label{sec:definitions}

We use \emph{trust} to refer to the trustor's subjective attitude toward, and willingness to rely on a system~\cite{lee2004trust,jacovi2021formalizing}. 
The primary trustors are clients, patients, and end users, although clinicians and institutions may also form trust judgments. We use \emph{trustworthiness} to refer to evaluable properties of the trustee: the AI system, interaction design, deployment workflow, or institution that may or may not deserve trust, including reliability, safety, privacy, fairness, transparency, accountability, and clinical appropriateness.

\revise{This distinction matters because trust can be warranted or unwarranted. Users may over-trust systems that communicate empathy, fluency, or human-likeness beyond actual capability, or under-trust safer systems whose safeguards are opaque. 
We therefore treat \emph{reliance} as a behavioral outcome related to, but not identical with trust. The goal is appropriate reliance: human trust should be calibrated to demonstrated interaction- and system-level capabilities rather than persuasive cues.}


\subsection{Stakeholder Landscape}
\label{sec:stakeholder}

\revise{We define stakeholder groups by their primary evaluation paradigms rather than by publication venues.
Venues were used only as search entry points in review process. The analytic categories below reflect what each community typically treats as evidence of trust or trustworthiness. \autoref{tab:stakeholder-mapping} summarizes these definitions and clarifies how client, patient, and end-user perspectives enter the framework.}


\begin{figure*}[!t]
\centering
\hspace*{-4.5pt}
\begin{tikzpicture}[
  font=\sffamily,
  title/.style={font=\bfseries\mainTitleSize},
  coltitle/.style={font=\bfseries\topTitleSize, text width=\cardW, align=center, inner sep=0pt, outer sep=0pt},
  stake/.style={font=\bfseries\stakeSize, text width=\stakeW, text=red!60!black, align=center, inner sep=0pt, outer sep=0pt},
  topbar/.style={line width=\topBarLineW, draw=black},
  arrow/.style={->, line width=\arrowLineW, draw=black},
  underline/.style={line width=\underlineLineW, draw=black},
  colbg/.style={
    rounded corners=\colBgCorner,
    fill=\colBgFill,
    draw=\colBgDraw,
    line width=\colBgLineW,
    inner xsep=\colBgPadX,
    inner ysep=\colBgPadY
  },
  card/.style={
    rounded corners=5pt,
    draw=black!18,
    line width=0.6pt,
    fill=white,
    font=\cardTextSize\sffamily,
    align=left,
    inner xsep=\cardInnerX,
    inner ysep=\cardInnerY,
    text width=\cardTextW,
    minimum width=\cardW,
    minimum height=6mm
  },
  cardTall/.style={card, minimum height=6mm},
  methodcard/.style={
    card,
    fill=\methodFill,
    draw=\methodDraw,
    line width=\methodLineW
  }
]

\pgfdeclarelayer{bg}
\pgfsetlayers{bg,main}

\def\mainTitleSize{\footnotesize}

\def\topBarLineW{0.55pt}
\def\arrowLineW{0.22pt}
\def\underlineLineW{0.75pt}

\def\arrowHeadLen{2.2mm}
\def\arrowHeadWid{1.9mm}
\tikzset{arrow/.style={-{Latex[length=\arrowHeadLen,width=\arrowHeadWid]}, line width=\arrowLineW, draw=black}}

\def\methodFill{blue!6}
\def\methodDraw{blue!20!black}
\def\methodLineW{0.6pt}

\def\colBgFill{black!3}
\def\colBgDraw{black!10}
\def\colBgLineW{0.4pt}
\def\colBgCorner{6pt}
\def\colBgPadX{2.0mm}
\def\colBgPadY{2.0mm}

\def\topTitleSize{\scriptsize} 
\def\stakeSize{\scriptsize}    

\def\cardTextSize{\scriptsize}

\def\cardWFixed{0pt} 

\def\colGapH{8mm} 

\def\fitEps{0.2pt} 
\ifdim\cardWFixed>0pt
  \pgfmathsetlengthmacro{\cardW}{\cardWFixed}
  \pgfmathsetlengthmacro{\colGapH}{0.5*((\linewidth - 2*\colBgPadX) - 3*\cardW)}
\else
  \pgfmathsetlengthmacro{\cardW}{((\linewidth - 2*\colBgPadX - 2*\colGapH)/3) - \fitEps}
\fi
\pgfmathsetlengthmacro{\colSep}{\cardW + \colGapH} 

\def\stakeExtraW{16pt}
\pgfmathsetlengthmacro{\stakeW}{\cardW + \stakeExtraW}

\def\cardInnerX{0.8mm} 
\def\cardInnerY{0.8mm} 
\pgfmathsetlengthmacro{\cardTextW}{\cardW - 2*\cardInnerX}

\def\yTitle{1.8mm}      
\def\yStake{-3.0mm}     
\def\yCards{-11.2mm}     
\def\cardGap{0.4mm}      
\def\underlineFrac{0.48} 
\pgfmathsetlengthmacro{\underlineHalf}{\underlineFrac*\stakeW}

\node[title] (T) at (0,0) {Three-Layer Trust Framework};

\node (C2) at (0,-12mm) {};
\node (C1) at ($(C2)+(-\colSep,0)$) {};
\node (C3) at ($(C2)+( \colSep,0)$) {};

\def\barY{8.2mm}           
\def\arrowDrop{4.0mm}      
\def\barPad{0mm}         

\coordinate (barL) at ($(C1)+( -\barPad, \barY)$);
\coordinate (barR) at ($(C3)+( +\barPad, \barY)$);
\draw[topbar] (barL) -- (barR);

\draw[arrow] ($(C1)+(0,\barY)$) -- ++(0,-\arrowDrop);
\draw[arrow] ($(C2)+(0,\barY)$) -- ++(0,-\arrowDrop);
\draw[arrow] ($(C3)+(0,\barY)$) -- ++(0,-\arrowDrop);

\node[coltitle] (Htitle) at ($(C1)+(0,\yTitle)$) {Human-Oriented Trust (\S\ref{human_oriented_trust})};
\node[stake] (Hstake) at ($(C1)+(0,\yStake)$) {{\footnotesize Common evidence}:\par Psychotherapy practitioners \& Users (\S\ref{psychotherapy});};
\draw[underline] ($(Hstake.south)+(-\underlineHalf,-2mm)$) -- ($(Hstake.south)+(\underlineHalf,-2mm)$);

\node[cardTall, anchor=north] (H1) at ($(C1)+(0,\yCards)$)
{\textbf{H1}: Perceived Trust \& Measures~\cite{CHI2021106700,LIU2022107026}};
\node[card, anchor=north] (H2) at ($(H1.south)+(0,-\cardGap)$) {\textbf{H2}: User Characteristics~\citep{KAUTTONEN2025,HUO2022107253}};
\node[card, anchor=north] (H3) at ($(H2.south)+(0,-\cardGap)$) {\textbf{H3}: System Characteristics~\citep{WU2023107614,LEICHTMANN2023107539}};
\node[methodcard, anchor=north] (Hmethod) at ($(H3.south)+(0,-\cardGap)$) {Evaluation \& Measures: \textbf{\textit{\underline{~\autoref{tab:literature_human}}}}};

\node[coltitle] (Ititle) at ($(C2)+(0,\yTitle)$) {Interaction-Oriented Trustworthiness (\S\ref{interaction_oriented_trust})};
\node[stake] (Istake) at ($(C2)+(0,\yStake)$) {{\footnotesize Common evidence}:\par HCI (\S\ref{hci}); Regulators (\S\ref{regulator});};
\draw[underline] ($(Istake.south)+(-\underlineHalf,-2mm)$) -- ($(Istake.south)+(\underlineHalf,-2mm)$);

\node[card, anchor=north] (I1) at ($(C2)+(0,\yCards)$) {\textbf{I1}: Competence~\citep{lee2025understanding,you2025tailored, zheng2025customizing}};
\node[card, anchor=north] (I2) at ($(I1.south)+(0,-\cardGap)$) {\textbf{I2}: Communication Style~\cite{namvarpour2024uncovering,ma2024evaluating}};
\node[card, anchor=north] (I3) at ($(I2.south)+(0,-\cardGap)$) {\textbf{I3}: Transparency~\cite{cao2025can}};
\node[card, anchor=north] (I4) at ($(I3.south)+(0,-\cardGap)$) {\textbf{I4}: Empathy \& Engagement~\citep{wang2025understanding,choi2025private}};
\node[card, anchor=north] (I5) at ($(I4.south)+(0,-\cardGap)$) {\textbf{I5}: Controllability~\cite{swinger2025there}};
\node[methodcard, anchor=north] (Imethod) at ($(I5.south)+(0,-\cardGap)$) {Method \& Evaluation: \textbf{\textit{\underline{~\autoref{tab:literature_interaction}}}}};

\node[coltitle] (Atitle) at ($(C3)+(0,\yTitle)$) {AI-Oriented Trustworthiness (\S\ref{ai_oriented_trust})};
\node[stake] (Astake) at ($(C3)+(0,\yStake)$) {{\footnotesize Common evidence}:\par AI/Safety/Security (\S\ref{ai};~\ref{safety})};
\draw[underline] ($(Astake.south)+(-\underlineHalf,-2mm)$) -- ($(Astake.south)+(\underlineHalf,-2mm)$);

\node[card, anchor=north] (A1) at ($(C3)+(0,\yCards)$) {\textbf{A1}: Reliability \& Robustness~\cite{kang-etal-2024-cure, dhuliawala-etal-2023-diachronic}};
\node[card, anchor=north] (A2) at ($(A1.south)+(0,-\cardGap)$) {\textbf{A2}: Safety \& Harm Prevention~\citep{hua-etal-2024-trustagent,baidal-etal-2025-guardians}};
\node[card, anchor=north] (A3) at ($(A2.south)+(0,-\cardGap)$) {\textbf{A3}: Privacy~\cite{shin-etal-2023-fedtherapist}};
\node[card, anchor=north] (A4) at ($(A3.south)+(0,-\cardGap)$) {\textbf{A4}: Explainability~\cite{yang-etal-2023-towards}};
\node[card, anchor=north] (A5) at ($(A4.south)+(0,-\cardGap)$) {\textbf{A5}: Fairness~\cite{gabriel-etal-2024-ai}};
\node[methodcard, anchor=north] (Amethod) at ($(A5.south)+(0,-\cardGap)$) {Method \& Evaluation: \textbf{\textit{\underline{~\autoref{tab:literature_ai}}}}};

\begin{pgfonlayer}{bg}
  \node[colbg, fit=(H1)(Hmethod)] {};
  \node[colbg, fit=(I1)(Imethod)] {};
  \node[colbg, fit=(A1)(Amethod)] {};
\end{pgfonlayer}

\end{tikzpicture}%
\vspace{-1.0mm}
\caption{The proposed Three-Layer Framework.
Each layer reflects dominant evaluation perspectives across stakeholders and summarizes key trust criteria with representative literature.
The evidence sources shown in each column are not exclusive to layer owners; individual literature may contribute evidence to multiple layers.
Examples of cross-layer evaluation signals and rubric are detailed in \rev{\autoref{tab:analytic-layer-assignments} and \autoref{tab:trust-rubric}}.
}
\label{fig:hai-trust}
\vspace{-0.4mm}
\end{figure*}


\subsubsection{Psychotherapy Practitioners}
\label{psychotherapy}
\revise{Psychotherapy practitioners include licensed therapists, clinical psychologists, psychiatrists acting in therapeutic roles, and mental-health researchers. 
This perspective treats user trust as clinically sensitive: trust is relational, dynamic, and depends on user characteristics, system cues, oversight, vulnerability, and perceived responsibility~\cite{CHI2021106700,KAUTTONEN2025,GILLE2025,Rai2025}. It is therefore concerned with calibrated reliance, not merely adoption or engagement.}
\revise{We distinguish mental-health-specific evidence from conversational support and counseling workflows~\cite{song2025typing,sun2025script,wester2024chatbot} from transferable foundations in broader high-risk decision and digital-health contexts~\cite{LEICHTMANN2023107539,MAYER2024108419,WOODCOCK2021}; the latter inform explanation, disclosure, and oversight mechanisms but do not substitute for mental-health evidence.}


\subsubsection{HCI Researchers}
\label{hci}
\revise{\Ac{HCI} researchers study how trust is communicated and negotiated for mental health support through interface design, conversational behavior, and lived experience. 
This perspective emphasizes transparency, empathy, engagement, and controllability: interaction should surface capability, limits, and safeguards in ways users can understand and act on~\citep{thieme2023designing,namvarpour2024uncovering,wang2025understanding,cao2025can}.}


\subsubsection{AI Researchers}
\label{ai}
\revise{AI and NLP researchers define which model, retrieval, data, and evaluation properties count as evidence of system-level trustworthiness. 
In mental health settings, this includes robustness, fairness, faithful grounding, RAG attribution, and evaluator reliability. Because LLM behavior can vary across runs and evaluation signals can be biased or unfaithful~\citep{cao2025writingstyle,liu2024robustir,wallat2025correctness}, model-level evidence is necessary but must be linked to interaction and human outcomes.}


\subsubsection{Safety and Security Researchers}
\label{safety}
Safety and security researchers ask whether systems remain safe under adversarial, high-risk, or privacy-sensitive conditions. Their evidence targets prompt injection, jailbreaks, backdoor triggers, memorization, and crisis stress tests~\citep{10.1145/3711896.3736561,hua-etal-2024-trustagent,10.5555/3666122.3667483,baidal-etal-2025-guardians,alghamdi-etal-2025-aratrust}. 
This perspective is important because mental health users disclose highly sensitive information and face severe consequences from unsafe advice or leakage.


\subsubsection{Regulators and Standards Associations}
\label{regulator}
Regulators and standards associations translate ethical principles into deployable requirements.
We review professional frameworks in U.S., EU~\citep{NIST2023,AMA2024,EPEUCO2024,AIHLEG2019,APA2025,Pillay2025}; despite different legal status, they converge on human-centricity, risk mitigation, privacy, bias auditing, accountability, and view that AI should augment rather than replace human clinical care.


\subsection{Three-Layer Framework}

\revise{Trust in mental health AI does not come from a single source. It is built through a calibration pathway: how the system works internally, how its capabilities and limits are communicated during interaction, and how users ultimately experience and rely on its support. To organize these views, we synthesize the following three-layer trust framework (Figure~\ref{framework};~\ref{fig:hai-trust}).}




\paragraph{Human-oriented trust.} 
\revise{Even if a system is technically robust and interactively competent, trust ultimately depends on whether users feel the system to be trustworthy and rely on it appropriately. 
Human-oriented trust captures users' subjective trust toward mental health AI systems. 
It concerns how trust is perceived, fostered, and calibrated by users, shaped by individual characteristics, expectations, prior experiences, vulnerability, and perceived stakes. This layer reflects trust as a psychological state rather than a system property.}



\paragraph{Interaction-oriented trustworthiness.} 
\revise{Once users begin interacting with systems, trustworthiness becomes user-visible. Interaction-oriented trustworthiness concerns how trust is mediated through conversational behavior, feedback, uncertainty communication, transparency cues, controllability, escalation, and safety mechanisms. This layer bridges system capability and user perception: it should communicate capability and limits, not simply make the system feel supportive.}



\paragraph{AI-oriented trustworthiness.} 
AI-oriented trustworthiness refers to system-level criteria defined at the model, data, evaluation, and infrastructure levels, including reliability, robustness, uncertainty awareness, fairness, privacy protection, evaluation validity, and avoidance of harmful or misleading outputs. These criteria are primarily specified and assessed by AI, NLP, and security research before and during deployment, and they constrain which interaction-level trustworthiness can be warranted. 
Failures at this layer, such as harmful generation, privacy leakage, or biased evaluation, may undermine trust regardless of interaction quality.




\revise{
Layer labels are assigned by evaluation target, not by keyword. 
Cross-layer assignments are expected: explainability may be evaluated as model faithfulness or explanation design; privacy as data protection or disclosure safety. 
Primary and secondary labels locate evidence across layers. 
(\autoref{tab:coding-examples} is coding examples and \autoref{tab:analytic-layer-assignments} is aggregate assignments.)}
\revise{The three-layer frame trust in MHS as calibration problem that should track interaction- and AI-oriented trustworthiness.
Taxonomy structures following review in \autoref{tab:trust-rubric}, which is evaluation aids, not certification or safety guarantee.}


\section{Human-Oriented Trust}
\label{human_oriented_trust}
\revise{Human-oriented trust is understood as a subjective psychological state that shapes users' willingness to adopt, disclose to, and continue using mental health AI systems.}
\revise{We organize human-oriented trust around three construct-level questions: what trust is, who trusts, and how trust is perceived through user-visible cues. This layer measures trust and reliance outcomes; it does not by itself establish whether the system deserves users' trust.}
Table~\ref{tab:literature_human} summarizes the criteria and measures.


\subsection{What Is Trust}

\paragraph{Scope.}
\revise{This dimension targets users' subjective trust judgments toward AI-powered mental health systems. At a basic level, trust reflects a user's confidence that an AI system is reliable, benevolent, and appropriate enough to rely on in a sensitive context. This aligns with trust theories such as the ability--benevolence--integrity model~\cite{abi} and MATCH framework~\cite{match}, which characterize trust as a belief about an agent's capability, dependability, and communicated trustworthiness cues.}
Prior work measures trust in two main ways. 
Some studies treat trust as a \emph{unidimensional} construct, measuring overall trust as a single judgment~\cite{WU2023107614,WOODCOCK2021,AOKI2021106572,KAUTTONEN2025,AKTAN2022107273,LIU2022107026,MAYER2024108419,HUO2022107253,Youn2021,ZHAO2025}. Others adopt a \emph{multidimensional} view, decomposing trust into components (e.g., perceived reliability, agency, and competence~\cite{CHI2021106700,LUETKELANFER2023,LEICHTMANN2023107539,GILLE2025,BRUNSWICKER2025108516,Rai2025}).

\paragraph{Evaluation.}
\revise{Trust is primarily measured through subjective assessments, while observed interaction behavior is better interpreted as evidence of reliance or use behavior unless paired with self-reported trust.}
Studies commonly employ validated scales capturing perceived agency~\cite{LUETKELANFER2023}, system reliability~\cite{CHI2021106700}, trust propensity~\cite{CHI2021106700}, or overall trust~\cite{WU2023107614,WOODCOCK2021,LIU2022107026,KAUTTONEN2025,AOKI2021106572,Youn2021}. 
\revise{Some work adopts single-item measures for direct trust assessment~\cite{WOODCOCK2021,AOKI2021106572,MAYER2024108419}. Qualitative approaches examine how users interpret and justify trust judgments~\cite{LUETKELANFER2023}. Behavioral indicators such as interaction patterns, decision choices, or continued use can reveal reliance and possible miscalibration~\cite{LEICHTMANN2023107539,BRUNSWICKER2025108516,MAYER2024108419,AKTAN2022107273}, but should not be treated as identical to subjective trust.}


\subsection{Who Trusts}
\paragraph{Scope.}
\revise{This dimension targets user-side characteristics that modulate trust formation. Prior work identifies attitudes toward AI, personality traits, familiarity, prior experience, and perceived social support as important determinants of trust in mental health contexts~\cite{KAUTTONEN2025,ZHAO2025,HUO2022107253}.}
Literacy level also matters: users with limited AI knowledge often exhibit greater reliance on AI~\cite{WOODCOCK2021}. 
Studies show that human-in-the-loop oversight increases trust in AI~\cite{MAYER2024108419,AOKI2021106572,LIU2022107026}, reflecting users’ preference for shared control in high-stakes contexts.

\paragraph{Evaluation.}
User characteristics are frequently measured using self-developed instruments tailored to demographic, socio-cultural factors~\cite{AKTAN2022107273,LEICHTMANN2023107539,ZHAO2025,HUO2022107253,WOODCOCK2021,LIU2022107026,KAUTTONEN2025}, while validated scales are more commonly used for stable traits such as personality or trust propensity.


\subsection{How Trust Is Perceived}

\paragraph{Scope.}
\revise{This dimension targets user-visible cues that shape human trust. While many system features may influence trust, psycho- and HCI-oriented studies place particular emphasis on anthropomorphism and explainability. In conversational mental health AI, anthropomorphism refers primarily to interactive cues, such as human-like language, empathy expressions, social presence, and emotional support, rather than embodiment or VR/AR immersion effects~\cite{wang2025understanding,cao2025can,LIU2022107026}.}
Explainability refers to system’s ability to communicate rationale behind its outputs and decisions in user-understandable ways~\cite{LEICHTMANN2023107539}.

\paragraph{Evaluation.}
\revise{
Anthropomorphism is commonly measured by validated scales~\cite{LIU2022107026}. 
Prior work distinguishes between superficial forms (e.g., appearance or communication style) and deeper forms involving moral agency and emotions~\cite{WU2023107614}. 
Studies often manipulate empathizing and systemizing behaviors before measuring perceived anthropomorphism~\cite{BRUNSWICKER2025108516}. 
Explainability is typically studied by varying clarity and form of system explanations and assessing user comprehension, confidence, and willingness to rely~\cite{LEICHTMANN2023107539,WOODCOCK2021}. 
Qualitative methods are frequently used to examine how users interpret these cues and whether the cues support calibrated trust~\cite{wang2025understanding,cao2025can,namvarpour2024uncovering,choi2025private}. These cues affect Layer~1 trust by signaling competence and emotional understanding, but they do not by themselves establish Layer~3 capability; this is the calibration risk foregrounded by the framework.}

\revise{Human-oriented trust thus captures how users perceive trust. 
These perceptions may diverge from interaction- and AI-level trustworthiness, which motivates evaluating calibrated trust across layers.}

\section{Interaction-Oriented Trustworthiness}
\label{interaction_oriented_trust}



\revise{Interaction-oriented trustworthiness asks whether system behavior supports appropriate trust during use.}
\revise{Unlike AI-oriented trustworthiness, it emerges in real time and is evaluated through observed conversations, interface affordances, and user comprehension. Its function is to translate system capability, uncertainty, limits, and safeguards into forms that users can understand and act on. Layer assignment is evidence-based: an NLP, AI, or health literature belongs here when its main evidence concerns user-centric behavior, control, or crisis routing.}
Table~\ref{tab:literature_interaction} summarizes the criteria and measures.


\subsection{Competence and Reliability}
\paragraph{Scope.}
\revise{Competence and reliability target whether MHS systems consistently provide accurate, contextually appropriate, and therapeutically meaningful responses during interaction.}
Prior work operationalizes this across application types: 
story-based interventions involve professional review of user-generated content for accuracy~\cite{sien2025gentel}.
Chatbots aim to recognize users’ emotions and deliver evidence-based interventions~\cite{wester2024chatbot}, while LLM agents generate responses aligned with psychological principles while avoiding ineffective or harmful advice \cite{song2025typing,sun2025script}.
Therapist-training systems assess competence via structured evaluations of treatments~\cite{swinger2025there}.

\paragraph{Methods and Evaluation.}
Literature studies competence through multiple ways. 
Qualitative interviews and analyses examine whether \ac{AI} provide useful and therapeutically relevant responses, while also documenting failures such as overly generic or inaccuracy~\cite{song2025typing}. 
Expertise-aligned generation ensures LLMs are controlled by expert-authored scripts, improving adherence to therapeutic principles~\cite{sun2025script}. 
In contrast, systems for therapist training rely on protocol-based assessments, such as time-stamped behavioral analysis and checklist-style scoring aligned with the clinic manuals~\cite{swinger2025there}.

\subsection{Conversational Safety and Controllability}
\paragraph{Scope.}
\revise{Conversational safety and controllability target how systems balance harm prevention with user agency during interaction. Safety mechanisms detect crisis, enforce boundaries, defer to human judgment in high-risk areas~\cite{song2025typing,sun2025script,swinger2025there}, while controllability ensures users can influence system behavior, such as selecting modules, controlling depth, or choosing system guidance level~\citep{sun2025script,wester2024chatbot,swinger2025there}.}

\paragraph{Methods and Evaluation.} 
To manage risk, studies commonly adopt modular or layered system designs. High-risk signals are routed to dedicated crisis modules or flagged for human intervention~\cite{sun2025script}. 
Autonomy-in-Middle architectures further restrict AI behavior to alerting while preserving human authority over decisions~\cite{swinger2025there}. 
Evaluation relies on user interviews and expert assessments that examine how systems handle emotionally nuanced and high-risk scenarios, often contrasting rigid filtering with more open-ended generative responses \cite{song2025typing}.
Crisis-oriented interaction tests should separate ordinary support, ambiguous distress, and self-harm by reporting over-refusals and escalation.


\subsection{Empathy and Engagement}
\paragraph{Scope.}
\revise{Empathy and engagement target whether users experience the interaction as emotionally appropriate without being misled about the system's actual understanding or clinical role.}
Literature identifies key tasks such as generating empathetic utterance, non-judgmental tone, and balancing therapeutic alliance with therapeutic structure~\citep{song2025typing,li-etal-2024-understanding-therapeutic,sun2025script}.

\paragraph{Methods and Evaluation.}
Systems operationalize these qualities through script- or protocol-aligned response generation \cite{sun2025script}, structured interaction designs (e.g., multi-choice interfaces or curated narratives) \citep{sien2025gentel,wester2024chatbot}, and expert ratings aligned with clinical competencies \cite{swinger2025there}. 
Evaluation commonly uses user self-report measures of engagement or emotional experience \citep{sien2025gentel,wester2024chatbot}, comparisons against rule-based or expert-aligned baselines \cite{sun2025script}, and expert coding of behaviors such as empathy and rapport in training contexts \cite{swinger2025there}.


\subsection{Transparency}
\paragraph{Scope.}
\revise{Transparency targets whether users can understand what the system is doing, including its limitations, decision boundaries, uncertainty, and the basis of its feedback.}
Literature highlights tasks such as clarifying system limitations, exposing decision logic at appropriate levels, and preventing misconceptions about human-like understanding or authority~\cite{song2025typing,sun2025script}.

\paragraph{Methods and Evaluation.}
Transparency is often implemented through explanations, annotations or labels that surface the rationale behind outputs \cite{sun2025script,swinger2025there}. 
User-facing artifacts such as journey maps or labeled strategies can improve interpretability in narrative-based and scripted settings \cite{sien2025gentel,sun2025script}. Evaluations consider whether transparency improves user comprehension, how it affects trust, and whether it introduces unintended interaction incentives (e.g., users adapting responses to satisfy the system rather than therapeutic goals) \cite{wester2024chatbot,swinger2025there}.

\subsection{Tensions and Trade-offs}
\paragraph{Scope.}
\revise{Interaction-oriented trustworthiness is inherently multi-dimensional. 
Competence, safety, empathy, transparency, and controllability cannot always be optimized independently. 
For example, strong refusal policies can reduce unsafe advice but also damage rapport; 
empathetic language can support disclosure but may exaggerate perceived understanding; 
detailed explanations can support informed use but increase cognitive burden; 
and user control can preserve autonomy while weakening adherence to therapeutic protocols.}

\paragraph{Methods and Evaluation.}
\revise{Interaction evaluation should report trade-offs rather than collapse criteria into single trust score.
Scenario-based tests can compare ordinary support, ambiguous distress, and crisis signals; expert review assesses therapeutic appropriateness and boundaries; user studies test if disclosures, explanations, and controls produce calibrated expectations.
This aligns with \autoref{tab:trust-rubric}: interactive cues should communicate capability and limits rather than merely persuade users to trust.}





\section{AI-Oriented Trustworthiness}
\label{ai_oriented_trust}

\revise{AI-oriented trustworthiness concerns whether a mental health AI system operates reliably, safely, and accountably at the model, data, retrieval, evaluation, and infrastructure levels. This layer is not venue-bound: HCI, digital-health, or clinical-adjacent studies are coded here when their main evidence targets model behavior, data governance, privacy, infrastructure, or evaluation validity.}
Table~\ref{tab:literature_ai} summarizes the criteria and measures.


\subsection{Reliability and Robustness}

\paragraph{Scope.} 
\revise{Reliability requires consistent model behavior across semantically similar or confusing inputs\rev{, and ability to know when it does not know: the system should flag uncertainty when it cannot respond safely}.
Robustness extends this requirement to long-tail, out-of-distribution, and adversarial conditions where inputs are indirect, emotionally charged, dialectal, or crafted to bypass safeguards~\citep{10.5555/3666122.3667483,alghamdi-etal-2025-aratrust}. 
Core tasks include risk, distress, and crisis identification, where false confidence can be especially harmful~\citep{yang-etal-2023-towards,cho-etal-2023-integrative}.}

\paragraph{Methods and Evaluation.} 
\revise{Methods span expertise alignment, uncertainty calibration, selective abstention, adversarial training, and multi-run evaluation for unstable LLM outputs~\citep{kang-etal-2024-cure,reuben-etal-2025-assessment,dhuliawala-etal-2023-diachronic,srivastava-etal-2024-knowledge,qiu-etal-2025-emoagent}. Evaluations include robustness audits, out-of-distribution stress tests, confidence calibration, refusal/abstention analysis, and error-cost analysis for crisis and high-risk scenarios~\citep{zhu-etal-2025-trust}.}
Stakeholders diverge: AI researchers prioritize consistency; practitioners prioritize avoiding risks; safety researchers prioritize resilience under manipulation \citep{cho-etal-2023-integrative, liu2023trustworthy}.


\subsection{Safety and Harm Prevention}
\paragraph{Scope.} 
\revise{Safety concerns the AI's capacity to avoid generating harmful, misleading, or clinically inappropriate content, especially when users disclose distress, self-harm risk, or crisis signals~\citep{cho-etal-2023-integrative,badawi2025position,na-etal-2025-survey}.}
Operationalization involves empathy, psychological-based attack, risk detection, content filters, and escalation triggers~\citep{chen-etal-2023-soulchat,hua-etal-2024-trustagent, 10.1145/3746252.3761164,zhang-etal-2024-psysafe}.

\paragraph{Methods and Evaluation.} 
\revise{Model-level safety integrates alignment, guardrails, policy-controlled refusals, escalation, and red-teaming~\citep{chen-etal-2023-soulchat,10.1145/3711896.3736561,hua-etal-2024-trustagent,liu2023trustworthy,qiu-etal-2025-emoagent}. 
Evaluation focuses on jailbreak resistance, crisis and ambiguous intent handling, false negatives that miss risk, and false positives that over-block supportive conversation~\citep{alghamdi-etal-2025-aratrust,baidal-etal-2025-guardians}.}
Stakeholder priorities diverge: 
safety researchers emphasize resistance, clinicians prioritize crisis recognition, and HCI researchers caution that overly restrictive refusals can undermine rapport~\citep{wang2025understanding,wester2024chatbot,cho-etal-2023-integrative}.

\subsection{Explainability}
\paragraph{Scope.} Explainability enables informed use of MHS systems by helping stakeholders understand how a system produces particular outputs~\cite{gollapalli-etal-2023-identifying,gabriel-etal-2024-ai}.

\paragraph{Methods and Evaluation.} Explainability can be supported through expert alignment, retrieval-augmented generation, tailored memory, and structured rationale generation~\cite{bi-etal-2025-magi,zhang-etal-2025-explainable,wang-etal-2025-annaagent,gollapalli-etal-2023-identifying}. However, explanation appearance is not sufficient evidence of trustworthiness: RAG citations may appear correct while failing to faithfully reflect evidence actually used by models~\cite{wallat2025correctness}. Evaluations should therefore inspect rationale quality, evidence faithfulness, criterion anchoring, and whether explanations are understandable to humans~\cite{zhai-etal-2025-mentalglm}. 
In mental health settings, the calibration risk is particularly acute: plausible-sounding explanations can create false reassurance about clinical relevance, motivating faithfulness audits separate from fluency measures.

\subsection{Fairness}
\paragraph{Scope.} 
Fairness addresses whether systems behave equitably across demographics, cultures, and languages~\cite{gabriel-etal-2024-ai,lissak-etal-2024-colorful}. In mental health, expressions of distress vary by culture, identity, dialect, and accessibility, so subgroup performance gaps can translate directly into uneven risk detection and uneven support quality.

\paragraph{Methods and Evaluation.} 
Fairness is assessed by comparing system behavior and outcomes across groups, stress-testing for distributional shift, and auditing biases in data, retrieval, generation, and evaluation pipelines~\cite{gabriel-etal-2024-ai,qi-etal-2025-kokorochat}. 
Mental-health-specific fairness evaluations should additionally consider matched-risk: (i) differential crisis-detection sensitivity across subgroups; (ii) variance in empathy and refusal behavior across demographic prompts; (iii) bias in LLM-as-judges that may systematically score certain dialects or affective styles as less coherent.


\subsection{Privacy and Data Protection}
\paragraph{Scope.} 
\revise{Privacy determines whether a system protects sensitive disclosures and prevents memorized information. In mental health contexts, perceived or actual exposure risk can undermine disclosure, reliance, and willingness to continue using MHS systems~\citep{baidal-etal-2025-guardians}. Many corpora derive from public forums or clinical notes with incomplete de-identification, making provenance and consent central to warranted trust~\citep{cho-etal-2023-integrative,qiu-lan-2025-psydial,cabrera-lozoya-etal-2025-synthetic}.}

\paragraph{Methods and Evaluation.} 
Privacy-preserving techniques include differential privacy, memorization audits, and federated learning~\citep{10.5555/3766078.3766387, shin-etal-2023-fedtherapist}. 
Evaluations from stakeholders diverge: regulators focus on compliance; security researchers map leakage vectors and extraction attacks~\citep{10.1145/3711896.3736561}; clinicians expect clinical-standard confidentiality; users equate privacy with safety~\citep{wang2025understanding}.

\revise{Across these criteria, AI-oriented trustworthiness provides necessary but incomplete evidence. Robust models, faithful explanations, and privacy-preserving pipelines must still be translated into interactions that users can understand, such as capability disclosure, escalation, and consent practices.}

\section{Discussion}
\revise{\textbf{Cross-layer calibration and key insights.} 
Trust fails when perceived trust and warranted trustworthiness diverge (see a worked example in Box~\ref{box:worked-example}).
The reviewed corpus shows why this matters, especially for mental health. 
Empathy and anthropomorphic language can increase disclosure and perceived understanding while also raising over-reliance if boundaries are unclear~\cite{song2025typing,sun2025script,BRUNSWICKER2025108516}. 
Crisis handling requires escalation, abstention, and false-negative analysis beyond ordinary user satisfaction~\cite{baidal-etal-2025-guardians,zhang-etal-2024-psysafe,alghamdi-etal-2025-aratrust}. 
Privacy functions as clinical confidentiality and emotional safety, not only data protection, because sensitive disclosures shape help-seeking~\cite{10.5555/3766078.3766387,baidal-etal-2025-guardians}. 
Thus, mental-health trust calibration cannot be reduced to aggregate model accuracy or general HAI trust measures (\autoref{tab:trust-rubric}; \autoref{tab:mental-health-specificity}).
}

\textbf{Implications for NLP research.} 
The objective for trustworthy mental health AI should be calibrated trust, not trust maximization: fluent, empathic, or personalized systems should clarify support boundaries rather than appear more capable than they are.
Model-level evaluations are partial evidence and should combine robustness, variability, privacy leakage, fairness, and faithful checks with interactions for ordinary support, ambiguous distress, crisis signals, privacy disclosure, and harmful-advice attempts. 
Human studies should then measure perceived trust, expectation alignment, and when to seek professional or emergency support. 
LLM-as-a-judge requires caution because evaluator bias can obscure clinically meaningful differences. 
LLM judge evaluations should report judge prompts, rubric anchors, bias pre-audits, and expert verification for crisis-sensitive subsets.

\vspace{1mm}
\begin{workedexamplebox}
  [box:worked-example]{Applying the Framework: A Worked Example}
For an empathy-tuned MHS LLM (SoulChat~\cite{chen-etal-2023-soulchat}), improved empathy ratings are insufficient for trustworthy support.
Our three-layer framework asks whether tuning preserves crisis handling, privacy, and robustness (AI-oriented trustworthiness); whether the interface communicates non-clinician status, boundaries, and escalation paths (Interaction-oriented trustworthiness); and whether users’ trust and reliance (Human-oriented trust) remain calibrated to those limits.
The key question is whether the system becomes safer or merely more trust-inducing.
\rev{The same check applies to unfaithful RAG citations~\cite{wallat2025correctness} and to over-refusal~\cite{wester2024chatbot}.}
\end{workedexamplebox}


\section{Conclusion}


\revise{
This survey synthesizes recent language-based mental health AI literature through a three-layer framework distinguishing human-oriented trust, interaction-oriented trustworthiness, and AI-oriented trustworthiness. We show that many trust failures arise from cross-layer and stakeholder misalignment, making calibrated trust more appropriate than trust maximization. The framework supports NLP evaluation, cross-disciplinary governance, and responsible deployment, but should be read as a survey-based evaluation lens rather than a clinical validation of AI mental health system.}




\section*{Limitations}
\label{sec:limitations}
\revise{This survey work has several limitations.}




\revise{First, our analysis is grounded in academic, peer-reviewed research literature. Real-world deployment practices, proprietary systems, patient advocacy perspectives, insurance and healthcare-system constraints, and the lived experiences of clinicians and users may reveal additional trust concerns that are not fully captured in published studies.}
Publicly reported incidents involving deployed mental-health or companion-chatbot systems, such as Koko’s AI-mediated support for young people~\footnote{https://kokocares.org/}, companion chatbots like Woebot~\footnote{https://woebothealth.com/} and Character.AI~\footnote{https://character.ai/}, illustrate why model-level evaluation is insufficient.
These cases point to cross-layer failures: unsafe or poorly governed system behavior, inadequate disclosure or boundary communication, inappropriate user reliance, and weak consent or safeguarding mechanisms. They are outside our analytic corpus and serve only as grey literature motivation. Still, they reinforce our central claim: trust failures in mental-health AI often emerge between layers, not within any single layer alone.

\revise{Second, despite broad coverage, trustworthy AI literature is rapidly evolving. Relevant work may have been missed or published outside our review window. 
Our 2021--2025 window was designed for the scoping corpus, while older foundational trust theory is used conceptually rather than exhaustively reviewed. 
The selected venues were intended to capture major publication channels for language-based mental-health AI trust evidence, but the review may miss relevant work in clinical informatics, psychiatry, broader digital medicine, and general machine-learning venues. Because the search is venue-stratified rather than database-exhaustive and query interfaces varied across sources, the review should be read as a PRISMA-guided scoping synthesis rather than an exhaustive clinical systematic review or meta-analysis.}

\revise{Third, our synthesis reflects how trust and trustworthiness are operationalized in existing research, which varies across disciplines. Some dimensions may appear more prominent because they are easier to benchmark or measure, rather than because they are intrinsically more important in practice. 
Coding was conducted as an author-team scoping synthesis using rules and examples reported in Appendix~\ref{appendix:methodology}. We therefore report coding rules, aggregate layer counts, representative examples, and criterion-level synthesis tables\rev{, plus a double-coded subset check,
}, 
rather than treating the review as a \rev{full} confirmatory double-coded systematic review.
Accordingly, layer counts should be interpreted as descriptive indicators of emphasis within scoped corpus rather than confirmatory estimates of full literature.
}

\revise{Lastly, our three-layer framework is conceptual rather than predictive: it organizes stakeholder perspectives and evaluation requirements but does not offer guarantees about system performance, clinical effectiveness, or user outcomes. Additional validation with clinicians, patient advocates, and deployment partners is needed before framework can be treated as an operational certification tool.}



\section*{Ethical Consideration}
\label{sec:ethics}

\revise{This work does not introduce new human-subject experiments or systems for mental health support. However, by synthesizing existing work, it aims to support more ethically grounded development, design, evaluation, and governance of trustworthy AI systems for mental health support. The survey should not be read as clinical validation of current AI mental health systems.}

\revise{There are several ethical concerns and potential risks to consider.}
\revise{First, AI systems in mental health raise heightened ethical risks due to user vulnerability and the sensitivity of personal disclosures. A primary concern is miscalibrated trust: systems may appear clinically competent or emotionally supportive beyond their actual capabilities, leading to over-reliance, delayed help-seeking, or inappropriate substitution for professional care. Anthropomorphic and empathetic interaction cues can further amplify this risk if not carefully constrained.}

\revise{Second, privacy and data protection are critical. Mental health AI often processes highly sensitive personal information, creating risks of data leakage, unintended memorization, secondary use without consent, or misuse. Safety validated in offline settings may not hold under real-world deployment.}

\revise{Third, bias and uneven performance across populations present additional ethical concerns. Models trained on limited or unrepresentative data may underperform for certain groups, dialects, cultures, or clinical presentations, potentially reinforcing existing mental health disparities.}

\revise{Lastly, survey and evaluation practices themselves pose ethical risks. Synthesizing existing literature can reproduce disciplinary blind spots if lived experience, non-English work, proprietary deployment failures, or clinical expertise are underrepresented. Over-reliance on automatic metrics or LLM-based evaluators can obscure failure modes and create a false sense of safety. Transparent reporting, human oversight, and clear communication of system limitations are therefore essential.}





\bibliography{main}


\newpage
\appendix


\section*{Appendix}

\section{AI Usage Disclosure}
\label{sec:appendix_ai_disclosure}
\label{sec:ai_disclosure}


We used AI tools in a supportive and limited role. 
Specifically, GPT-5 was used for language editing (e.g., improving clarity and conciseness). All findings and figures are based on our own data and results. 
The survey scoping, literature review, data analysis, and final content were conducted and verified by the authors.

\section{Methodology of Literature Review}
\label{appendix:methodology}

To provide a transparency for a cross-disciplinary survey, we conducted a PRISMA-guided scoping review~\cite{PRISMA}. 
We use PRISMA principles to structure screening and reporting, but the study is not a database-exhaustive clinical systematic review or meta-analysis. Instead, it targets recent peer-reviewed work on language-based AI systems for mental health support and uses an auditable screening and coding procedure to synthesize how trust and trustworthiness are operationalized across research communities.

\begin{figure}[H]
	\centering
	\includegraphics[width=0.478\textwidth]{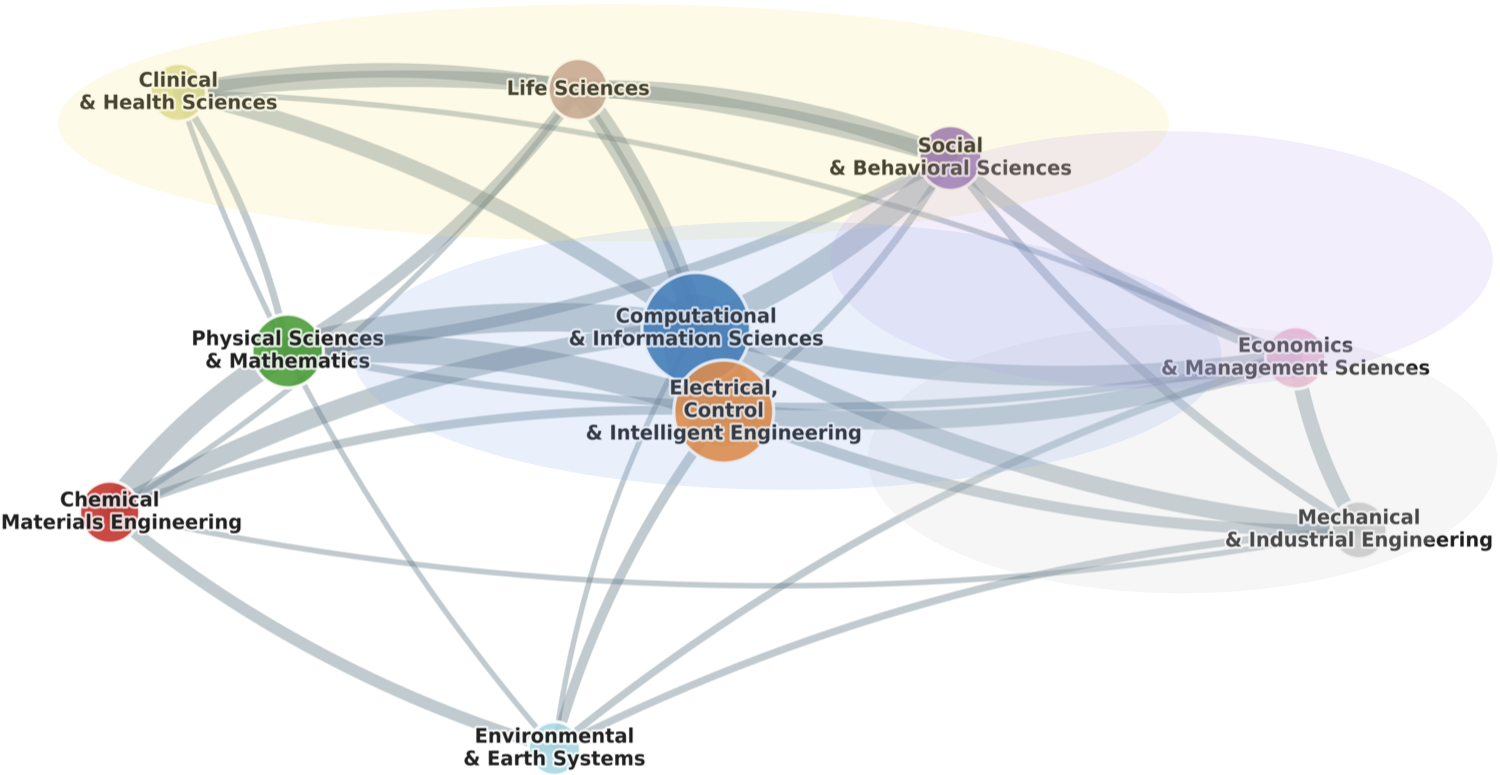}
	\vspace{-4.6mm}
	\caption{\revise{Discipline network from a broad 1,706-paper mapping corpus (2021--2025) related to trust in mental health AI. The network visualizes the interdisciplinary landscape. Visualization source code is released in an \href{https://anonymous.4open.science/status/Trust-Survey-F74F}{anonymous repository}.}}
	\label{fig:example}
	\vspace{-3.0mm}
\end{figure}

\begin{figure*}[!tbp]
\centering
\small
\begin{tikzpicture}[
	font=\tiny,
	mainbox/.style={
		draw=black!60,
		rounded corners=2pt,
		align=center,
		inner sep=4pt,
		text width=0.48\textwidth
	},
	sidebox/.style={
		draw=black!45,
		rounded corners=2pt,
		align=flush left,
		inner sep=4pt,
		text width=0.34\textwidth,
		execute at begin node=\setlength{\baselineskip}{7.2pt}
	},
	arrow/.style={
		-{Latex[length=2.1mm,width=1.8mm]},
		line width=0.45pt,
		draw=black!70
	}
]

\def\mainX{0}

\def\sideX{8.60}

\def\sideArrowLen{1.0}

\def\rightArrowGap{0.20}


\node[mainbox] (identified) at (\mainX,0) {
\textbf{Records identified through stratified search:} $n = 735$\\
Digital health / behavioral science: $252$; NLP / AI / security: $106$\\
HCI / human-centered computing: $377$
};

\node[mainbox] (screened) at (\mainX,-1.55) {
\textbf{Records after title/abstract screening:} $n = 218$
};

\node[sidebox] (excludedTitle) at (\sideX,-1.35) {
\textbf{Excluded after title/abstract:} $n = 517$\\
Digital health / behavioral science: $174$\\
NLP / AI / security: $40$\\
HCI / human-centered: $303$
};

\node[mainbox] (fulltext) at (\mainX,-3.25) {
\textbf{Full-text records assessed for eligibility:} $n = 218$
};

\node[sidebox] (excludedFull) at (\sideX,-3.95) {
\textbf{Excluded after full-text review:} $n = 145$\\
Digital health / behavioral science: $62$\\
NLP / AI / security: $25$\\
HCI / human-centered: $58$\\
\textit{Reason categories:}\\
non-language or non-MHS focus;\\
no trust construct;\\
no design/evaluation evidence;\\
survey/review or non-peer-reviewed item
};

\node[mainbox] (eligible) at (\mainX,-5.35) {
\textbf{Eligible records before cross-stratum de-duplication:} $n = 73$
};

\node[mainbox] (unique) at (\mainX,-6.75) {
\textbf{Unique papers included in qualitative synthesis:} $n = 61$
};

\node[mainbox] (assignments) at (\mainX,-8.25) {
\textbf{Analytic layer assignments after coding of the 61 unique papers:}\\
Human-oriented trust: $16$; Interaction-oriented trustworthiness: $16$; AI-oriented trustworthiness: $41$\\
\textit{Sum} $=73$ because some papers contribute to multiple layers
};


\draw[arrow] (identified) -- (screened);
\draw[arrow] (screened) -- (fulltext);
\draw[arrow] (fulltext) -- (eligible);
\draw[arrow] (eligible) -- (unique);
\draw[arrow] (unique) -- (assignments);

\draw[arrow]
	(identified.east) -- ++(\sideArrowLen,0)
	|- ($(excludedTitle.west)+(-\rightArrowGap,0)$);

\draw[arrow]
	(fulltext.east) -- ++(\sideArrowLen,0)
	|- ($(excludedFull.west)+(-\rightArrowGap,0)$);

\end{tikzpicture}
\caption{\revise{PRISMA-style flow for the PRISMA-guided scoping review. Search strata were used only for retrieval coverage and were not analytic layer labels. The full-text exclusion box lists eligibility categories applied during review. The final corpus contains 61 unique papers; layer assignments are reported as 16/16/41 because some papers contributed evidence to more than one layer. 
}}
\label{fig:prisma-flow}
\end{figure*}
\begin{table*}[!tbp]
\centering
\small
\renewcommand{\arraystretch}{1.16}
\setlength{\tabcolsep}{5pt}
\begin{tabularx}{\textwidth}{@{}p{0.20\textwidth}p{0.24\textwidth}X@{}}
\toprule
\textbf{\revise{Search entry stratum}} & \textbf{\revise{Venues / sources}} & \textbf{\revise{Purpose in retrieval}} \\
\midrule
\revise{Digital health / behavioral science} & \revise{JMIR, CHB} & \revise{Capture health, disclosure, user trust, perceived competence, and clinical-adjacent evidence relevant to mental health support.} \\
\revise{NLP / AI / security} & \revise{ACL, EMNLP, NAACL, COLING, IJCAI, USENIX} & \revise{Capture model, retrieval, robustness, safety, privacy, fairness, and evaluation evidence from language-based AI systems.} \\
\revise{HCI / human-centered computing} & \revise{ACM CHI, ACM CSCW, TOCHI, IJHCS} & \revise{Capture interaction design, lived experience, transparency, controllability, crisis routing, and conversational safety evidence.} \\
\bottomrule
\end{tabularx}
\caption{\revise{Search entry strata used for retrieval coverage. It reports where papers were searched from. These strata are not analytic layer labels and do not determine how papers are assigned to the three-layer framework.}}
\label{tab:corpus-overview}
\end{table*}

\begin{table*}[!tbp]
\centering
\small
\renewcommand{\arraystretch}{1.16}
\setlength{\tabcolsep}{5pt}
\begin{tabularx}{\textwidth}{@{}p{0.23\textwidth}p{0.13\textwidth}p{0.30\textwidth}X@{}}
\toprule
\textbf{\revise{Analytic layer}} & \textbf{\revise{Assignments}} & \textbf{\revise{Assignment criterion}} & \textbf{\revise{Example constructs}} \\
\midrule
\revise{Human-oriented trust} & \revise{$n=16$} & \revise{The primary evaluation target is a user's subjective trust state, expectation, or reliance outcome.} & \revise{Perceived trust, trust propensity, disclosure, user characteristics, reliance behavior.} \\
\revise{Interaction-oriented trustworthiness} & \revise{$n=16$} & \revise{The primary evaluation target is user-visible system behavior or interaction design.} & \revise{Empathy, transparency, controllability, crisis routing, uncertainty communication, boundary setting.} \\
\revise{AI-oriented trustworthiness} & \revise{$n=41$} & \revise{The primary evaluation target is a model-, data-, retrieval-, evaluation-, privacy-, or infrastructure-level property.} & \revise{Robustness, safety, privacy, fairness, explanation faithfulness, evaluator reliability.} \\
\bottomrule
\end{tabularx}
\caption{
Analytic layer assignments are produced after coding. The final corpus contains 61 unique papers; assignments sum to 73 because secondary labels were used when a paper provided substantive evidence for more than one layer. 
This corresponds to 12 additional layer assignments beyond the 61 unique papers, not 12 additional papers.
These counts describe the distribution of evidence within our scoped corpus rather than field-wide prevalence estimates; they are used to identify recurring emphases and gaps, not to make statistical claims about the literature.
}
\label{tab:analytic-layer-assignments}
\end{table*}
\begin{table*}[!tbp]
\centering
\scriptsize
\renewcommand{\arraystretch}{1.18}
\setlength{\tabcolsep}{4pt}
\begin{tabularx}{\textwidth}{@{}p{0.15\textwidth}p{0.18\textwidth}p{0.23\textwidth}p{0.19\textwidth}X@{}}
\toprule
\textbf{\revise{Layer}} & \textbf{\revise{Core evaluation question}} & \textbf{\revise{Typical methods}} & \textbf{\revise{Risk if omitted}} & \textbf{\revise{Implication for NLP systems}} \\
\midrule
\revise{Human-oriented trust} & \revise{Do users' trust and reliance match what the system can safely provide?} & \revise{Validated trust scales, interviews, reliance behavior, expectation checks, longitudinal use and disengagement indicators.} & \revise{Persuasive systems may create over-trust, while opaque but safer systems may create under-trust or abandonment.} & \revise{Report user-facing outcomes, reliance, and expectation calibration rather than only model scores.} \\
\revise{Interaction-oriented trustworthiness} & \revise{Do interaction behaviors communicate capability, limits, uncertainty, and safety boundaries appropriately?} & \revise{Scenario-based dialogue tests, expert review, crisis-routing evaluation, transparency and controllability studies, user comprehension checks.} & \revise{Empathy, fluency, or anthropomorphic cues may inflate trust without corresponding capability; excessive refusals may damage rapport.} & \revise{Evaluate responses in realistic mental-health scenarios, including uncertainty communication and boundary setting.} \\
\revise{AI-oriented trustworthiness} & \revise{Does the underlying model or pipeline deserve trust under normal, long-tail, and adversarial conditions?} & \revise{Robustness and fairness audits, calibration and abstention tests, privacy leakage tests, RAG faithfulness checks, LLM-evaluator bias audits.} & \revise{Benchmark performance may overstate safety, faithfulness, privacy, or clinical reliability.} & \revise{Treat RAG, LLM-as-a-judge, and aggregate metrics as partial evidence that must be linked to interaction and user outcomes.} \\
\bottomrule
\end{tabularx}
\caption{
Integrated trust evaluation rubric. The rubric synthesizes reviewed practices and identifies evaluation gaps so that evaluations target calibrated trust rather than trust maximization. It is an evaluation checklist, not an empirically validated scoring instrument, certification instrument, or guarantee of safety.
\rev{In practice, calibration can be measured as the gap between user trust or reliance and demonstrated capability, e.g., rates of following incorrect advice (over-reliance) and rejecting correct support (under-reliance) in scenario tests, making the rubric usable as a three-layer report card.}
A minimal \emph{calibrated-trust} report should include one signal from each layer:
\emph{L1:} humans perceived trust or expectation-alignment measure;
\emph{L2:} boundary comprehension, uncertainty disclosure, and escalation behavior under ordinary, ambiguous, and crisis-like scenarios;
\emph{L3:} robustness/abstention, privacy leakage, fairness, and faithful-grounding checks.
Claims of trustworthy support should report cross-layer consistency rather than a single trust score or metric.
}
\label{tab:trust-rubric}
\vspace{1.8mm}
\end{table*}

\begin{table*}[!tbp]
\centering
\scriptsize
\renewcommand{\arraystretch}{1.16}
\setlength{\tabcolsep}{3.5pt}
\begin{tabularx}{\textwidth}{@{}p{0.18\textwidth}p{0.18\textwidth}p{0.19\textwidth}p{0.18\textwidth}X@{}}
\toprule
\textbf{\revise{Representative evidence}} & \textbf{\revise{Primary construct}} & \textbf{\revise{Evaluation target}} & \textbf{\revise{Layer label rule}} & \textbf{\revise{Why not venue-based}} \\
\midrule
\revise{Trust scales and user studies~\cite{CHI2021106700,KAUTTONEN2025,LIU2022107026}} & \revise{Perceived trust, trust propensity, user characteristics} & \revise{Subjective trust response and reliance tendency} & \revise{Primary: human-oriented trust} & \revise{The label follows the measured construct. A user study in any venue is human-oriented when its central evidence is users' trust or reliance.} \\
\revise{Scripted or protocol-aligned chatbot interactions~\cite{song2025typing,sun2025script,wester2024chatbot}} & \revise{Empathy, controllability, transparency, therapeutic alignment} & \revise{Observable conversational behavior and user-facing controls} & \revise{Primary: interaction-oriented trustworthiness} & \revise{The label follows the interaction evidence, regardless of whether the paper appears in HCI, AI, or health venues.} \\
\revise{ACL/NAACL user-facing counseling studies~\cite{lissak-etal-2024-colorful,qi-etal-2025-kokorochat}} & \revise{Subgroup support quality, counseling dialogue quality, client feedback} & \revise{User-visible support behavior and subgroup-sensitive response quality} & \revise{Primary: interaction-oriented when evidence concerns support behavior; secondary: AI-oriented when it audits model or data properties} & \revise{An ACL/NAACL paper is not automatically AI-oriented; the label follows the evaluation target, not the venue.} \\
\revise{RAG and explanation faithfulness studies~\cite{wallat2025correctness,zhang-etal-2025-explainable,gollapalli-etal-2023-identifying}} & \revise{Faithfulness, attribution quality, rationale inspection} & \revise{Model or pipeline evidence for whether explanations deserve trust} & \revise{Primary: AI-oriented trustworthiness; secondary: interaction-oriented when explanations are user-facing} & \revise{A technical paper can receive an interaction label only when it evaluates how explanations are exposed to or interpreted by users.} \\
\revise{Safety, jailbreak, and crisis-handling evaluations~\cite{hua-etal-2024-trustagent,baidal-etal-2025-guardians,alghamdi-etal-2025-aratrust}} & \revise{Harm prevention, crisis escalation, adversarial robustness} & \revise{Worst-case system behavior and high-risk interaction outcomes} & \revise{Primary: AI-oriented trustworthiness; secondary: interaction-oriented for crisis routing} & \revise{The primary label reflects system safety evidence; a secondary label is added only when user-visible routing or escalation behavior is substantively evaluated.} \\
\revise{Privacy-preserving mental-health AI~\cite{shin-etal-2023-fedtherapist,10.5555/3766078.3766387}} & \revise{Privacy, data protection, memorization risk} & \revise{Sensitive-data handling and leakage resistance} & \revise{Primary: AI-oriented trustworthiness} & \revise{The evidence concerns data and infrastructure protection, which constrains warranted trust even when no user study is present.} \\
\bottomrule
\end{tabularx}
\caption{\revise{Representative coding examples showing how papers were mapped from trust-related constructs to framework layers. Primary labels follow the main evaluation target; secondary labels are used only when a paper provides substantive evidence for another layer. These examples document the mapping logic for a PRISMA-guided scoping synthesis and are not intended as a confirmatory inter-rater reliability report.}}
\label{tab:coding-examples}
\end{table*}
\begin{table*}[!tbp]
\centering
\scriptsize
\renewcommand{\arraystretch}{1.18}
\setlength{\tabcolsep}{4pt}
\begin{tabularx}{\textwidth}{@{}p{0.16\textwidth}p{0.22\textwidth}p{0.22\textwidth}X@{}}
\toprule
\textbf{\revise{Stakeholder / perspective}} & \textbf{\revise{Definition in this survey}} & \textbf{\revise{Typical evidence}} & \textbf{\revise{How users or clients enter the framework}} \\
\midrule
\revise{Clients, patients, and end users} & \revise{Primary trustors whose subjective trust, reliance, disclosure, and help-seeking outcomes are affected by mental health AI.} & \revise{Self-report scales, interviews, participatory design, preference tasks, usage and reliance indicators.} & \revise{Represented centrally in the human-oriented layer rather than treated as a publication-venue category.} \\
\revise{Psychotherapy practitioners and clinical researchers} & \revise{Licensed therapists, clinical psychologists, psychiatrists acting in therapeutic roles, and mental-health researchers concerned with therapeutic fidelity and client safety.} & \revise{Therapeutic alliance, clinical appropriateness, crisis safety, user vulnerability, and human oversight evidence.} & \revise{Bring client needs, therapeutic norms, confidentiality, and responsibility into the interpretation of warranted trust.} \\
\revise{HCI researchers} & \revise{Researchers who study user-centered design, interaction patterns, lived experience, and empirical use of AI-mediated mental-health support.} & \revise{User studies, interviews, interface evaluations, design probes, scenario studies, and interaction analyses.} & \revise{Translate lived-experience evidence into interaction-level criteria such as transparency, controllability, empathy, and boundary communication.} \\
\revise{AI/NLP researchers} & \revise{Researchers who develop or evaluate model, retrieval, generation, classification, and automatic evaluation methods.} & \revise{Benchmarks, robustness tests, RAG faithfulness analyses, LLM-as-a-judge evaluations, calibration, and fairness audits.} & \revise{Provide evidence about whether system capabilities can justify the level of trust encouraged by user-facing interaction.} \\
\revise{Safety and security researchers} & \revise{Researchers who examine adversarial misuse, jailbreaks, privacy leakage, memorization, and worst-case failures.} & \revise{Red-teaming, jailbreak audits, membership inference, extraction attacks, and crisis scenario stress tests.} & \revise{Identify failure modes that are especially consequential when vulnerable users disclose sensitive information.} \\
\revise{Regulators and standards bodies} & \revise{Institutions and professional bodies that formalize minimum requirements for risk management, accountability, privacy, and oversight.} & \revise{Regulatory frameworks, professional ethics guidance, risk management standards, and accountability principles.} & \revise{Define external constraints under which user trust may be considered warranted or institutionally acceptable.} \\
\bottomrule
\end{tabularx}
\caption{\revise{Stakeholder definitions used in this survey. Categories are based on evaluation paradigms and roles in trust calibration, not on publication venue boundaries.}}
\label{tab:stakeholder-mapping}
\end{table*}

\begin{table*}[!tbp]
\centering
\scriptsize
\renewcommand{\arraystretch}{1.16}
\setlength{\tabcolsep}{4pt}
\begin{tabularx}{\textwidth}{@{}p{0.16\textwidth}p{0.29\textwidth}p{0.25\textwidth}X@{}}
\toprule
\textbf{\revise{General HAI issue}} & \textbf{\revise{Mental-health-specific pressure}} & \textbf{\revise{Representative reviewed evidence}} & \textbf{\revise{Consequence for this framework}} \\
\midrule
\revise{Reliance} & \revise{Users may be distressed, vulnerable, or unsure whether to seek professional help.} & \revise{User trust and reliance studies~\cite{CHI2021106700,WOODCOCK2021,MAYER2024108419}; crisis-oriented safety work~\cite{hua-etal-2024-trustagent,baidal-etal-2025-guardians}.} & \revise{Evaluate appropriate reliance, help-seeking, and disengagement, not only adoption or engagement.} \\
\revise{Empathy} & \revise{Warm language can invite disclosure while also inflating perceived competence or clinical authority.} & \revise{Anthropomorphism and empathy studies~\cite{WU2023107614,LIU2022107026,BRUNSWICKER2025108516}; scripted support systems~\cite{sun2025script,song2025typing}.} & \revise{Test whether empathy calibrates expectations or exaggerates what the system can safely provide.} \\
\revise{Transparency} & \revise{Users may mistake AI support for therapy, diagnosis, or professional oversight.} & \revise{Explanation and transparency studies~\cite{LEICHTMANN2023107539,WOODCOCK2021,sun2025script,wester2024chatbot}.} & \revise{Require boundary disclosure, uncertainty communication, and clear escalation pathways.} \\
\revise{Privacy} & \revise{Disclosures may resemble clinical-confidential information and include trauma, stigma, or crisis details.} & \revise{Privacy and leakage evaluations~\cite{shin-etal-2023-fedtherapist,10.5555/3766078.3766387,10.1145/3711896.3736561}.} & \revise{Evaluate consent, retention, leakage, memorization, and perceived privacy safety.} \\
\revise{Robustness} & \revise{Distress and self-harm cues can be ambiguous, indirect, culturally varied, or adversarially phrased.} & \revise{Robustness, calibration, and adversarial tests~\cite{kang-etal-2024-cure,dhuliawala-etal-2023-diachronic,alghamdi-etal-2025-aratrust}.} & \revise{Use scenario stress tests for ordinary support, ambiguous distress, and crisis cases.} \\
\revise{Fairness} & \revise{Expressions of distress vary by culture, language, identity, and access to care.} & \revise{Fairness and subgroup analyses~\cite{gabriel-etal-2024-ai,lissak-etal-2024-colorful,qi-etal-2025-kokorochat}.} & \revise{Audit differential risk detection, support quality, refusal behavior, and escalation across groups.} \\
\revise{Explanation} & \revise{Plausible explanations may create false reassurance in high-stakes support conversations.} & \revise{RAG and explanation-faithfulness evaluations~\cite{wallat2025correctness,zhang-etal-2025-explainable,gollapalli-etal-2023-identifying}.} & \revise{Evaluate faithfulness, user comprehension, and clinical relevance rather than explanation fluency alone.} \\
\bottomrule
\end{tabularx}
\caption{\revise{How mental health support changes general human-AI trust evaluation. The table links domain-specific pressures to representative evidence from the reviewed corpus and explains why this survey emphasizes trust calibration across human, interaction, and AI-oriented layers rather than perceived trust alone.} \rev{These pressures operate at the framework level, not only in the analysis details: in mental health, the trust relationship is itself part of the intervention, so miscalibrated trust directly harms disclosure, help-seeking, and crisis safety rather than only task accuracy. This separates our aim from general (X)AI trust and reliance surveys~\citep{visser2025trust,match}, which target general task settings.}}
\label{tab:mental-health-specificity}
\end{table*}

\paragraph{Scope and Search Strategy.}
\revise{The scoping corpus covers papers published between Jan 2021 and Nov 2025. 
The date range applies only to corpus selection: it was chosen to capture the recent shift toward LLM-powered conversational AI and rapidly changing evaluation practices in mental health AI. Foundational trust theory before 2021 is still used for conceptual grounding throughout the paper.}

\revise{The 1,706-paper corpus in \autoref{fig:example} is a broader bibliometric mapping corpus used to visualize disciplinary connections around trust and mental health AI. It is not identical to the analytic corpus used for coding. The PRISMA-guided scoping review below uses a narrower stratified search of 735 records and retains 61 unique papers that meet the language-based mental-health AI and trust-related inclusion criteria. Thus, \autoref{fig:example} motivates the interdisciplinary landscape, whereas this appendix defines the corpus used for synthesis.}

\revise{We use mental health support as an umbrella term for screening, psychoeducation, conversational support, coaching, therapy-related training, and psychological self-help. These applications are not clinically equivalent: low-risk psychoeducation primarily requires transparency and privacy safeguards, whereas crisis-adjacent or psychotherapy-adjacent systems require stronger escalation, abstention, human oversight, and accountability.}

\revise{We searched across three disciplinary strata motivated by the network in \autoref{fig:example}: digital health and behavioral science outlets (JMIR, CHB), NLP/AI/security venues (ACL, EMNLP, NAACL, COLING, IJCAI, USENIX), and HCI venues and journals (ACM CHI, ACM CSCW, TOCHI, IJHCS). These strata were search entry points for retrieval coverage, not analytic categories or proxies for trust layers (\autoref{tab:corpus-overview}). Search terms combined mental-health concepts with AI-system and trust-related concepts, including variants of \textit{AI}, \textit{LLM}, \textit{dialogue system}, \textit{chatbot}, \textit{mental health}, \textit{psychological support}, \textit{safety}, \textit{trust}, \textit{trustworthiness}, \textit{privacy}, \textit{reliability}, and \textit{explainability}. Searches were conducted over titles, abstracts, and main text when available, excluding reference sections.}

\paragraph{Search Implementation and Reproducibility.}
\revise{Searches were implemented as venue-stratified queries rather than a single database-wide query. We searched title, abstract, and available full-text fields, excluding reference sections. The query templates combined three concept groups: (mental health OR psychological support OR therapy OR counseling OR wellbeing) AND (AI OR LLM OR chatbot OR dialogue system OR conversational agent) AND (trust OR trustworthiness OR safety OR privacy OR reliability OR robustness OR fairness OR explainability OR transparency). Because search interfaces varied across proceedings sites, publisher portals, and manual venue browsing, query implementation was standardized at the concept-group level rather than as byte-identical query strings across all sources. We document the search strata, query templates, eligibility criteria, screening counts, final corpus size, coding rules, and representative coding examples to support a PRISMA-guided scoping synthesis. This venue-stratified design improves cross-disciplinary coverage but may miss relevant work in clinical informatics, psychiatry, broader digital medicine, AI ethics, and general machine-learning venues.}

\paragraph{Inclusion and Exclusion Criteria.}
\revise{As shown in Figure~\ref{fig:prisma-flow}, we included peer-reviewed papers that met all three criteria: (1) the paper studied, evaluated, or analyzed a language-based AI system or AI-mediated workflow for mental health support, screening, coaching, therapy-related training, or psychological self-help; (2) the paper explicitly addressed at least one trust-related construct or criterion, even if it did not use the word ``trust'' directly; and (3) the paper provided empirical findings, system design, evaluation methodology, technical analysis, or clinically relevant design evidence. Trust-related constructs included user trust, reliance, therapeutic alliance, perceived competence, interaction quality, empathy, transparency, controllability, reliability, robustness, fairness, privacy, safety, crisis handling, and evaluation validity.}

\revise{We excluded papers that focused only on non-health domains, mental health without AI, non-language or sensor-only interventions outside the scope of this survey, purely theoretical arguments without design or evaluation evidence, tutorials, extended abstracts, non-peer-reviewed articles, and prior surveys. Prior surveys were excluded from the scoping corpus but used as related work when they helped position our contribution. Broader health or digital-health studies were included only when they directly informed transferable mechanisms relevant to mental health AI trust, such as sensitive disclosure, user vulnerability, clinician accountability, or privacy expectations.}

\paragraph{Screening and Selection.}
\revise{Screening followed a multi-stage process summarized in \autoref{fig:prisma-flow}. First, titles and abstracts were screened to remove records that clearly lacked an AI system, mental-health relevance, or any trust-related criterion. Second, full texts were reviewed against the inclusion and exclusion criteria. Full-text exclusions were applied for reasons such as non-language-based AI, no mental-health support focus, no trust-related construct, survey or review status, non-peer-reviewed or tutorial format, purely theoretical scope, or absence of design/evaluation evidence. We report these exclusion categories to clarify eligibility decisions, while treating the synthesis as a scoping review rather than a reason-by-reason exclusion audit. Third, retained papers were de-duplicated across strata and checked for cross-layer relevance. The final corpus contains 61 unique papers. Because a single paper can contribute evidence to more than one layer, the layer-level assignments are reported as Human-oriented $=16$, Interaction-oriented $=16$, and AI-oriented $=41$ (sum $=73$), corresponding to 12 additional layer assignments beyond the 61 unique papers (\autoref{tab:analytic-layer-assignments}).}
This distribution is itself a finding. The AI-oriented evidence base is approximately 2.5× the human-oriented base, despite the centrality of user trust in mental health support. Even when we restrict to papers published in 2023–2025 (the post-LLM-conversational-MHS period), the imbalance persists (AI-oriented = 24, Human-oriented = 9, Interaction-oriented = 11). This asymmetry is not solely a venue-selection artifact: digital-health journals (JMIR, CHB) were searched specifically to capture human-oriented evidence, yet AI/NLP venues yielded substantially more language-based MHS trust studies. 
We interpret this distribution as evidence of the calibration gap our framework articulates: the community has invested heavily in measuring system-level properties (robustness, fairness, privacy) while comparatively underinvesting in measuring whether users' subjective trust and reliance track those properties in mental health contexts.

\begin{table*}[!tbp]
\centering
\scriptsize
\renewcommand{\arraystretch}{1.36}
\resizebox{\textwidth}{!}{
\setlength{\tabcolsep}{3pt} 
\begin{tabular}{@{}p{3.5cm}p{2.2cm}p{0.5cm}p{8cm}@{}}
\toprule
\textbf{Document} & \textbf{Agency} & \textbf{Year} & \textbf{Key Focus of Trustworthiness}   \\ \midrule
NIST AI Risk Management Framework (AI RMF 1.0)~\cite{NIST2023}
& National Institute of Standards and Technology
& 2023          
& Focuses on seven characteristics of trustworthy AI to manage risks and promote responsible use: Valid and Reliable, Safe, Secure and Resilient, Accountable and Transparent, Explainable and Interpretable, Privacy-Enhanced, and Fair with Harmful Bias Managed. Accountability and Transparency are emphasized as relating to all other characteristics and internal processes.      
\\
Artificial Intelligence Act (AI Act)~\cite{EPEUCO2024}
& European Parliament and Council 
& 2024          
& Promotes human-centric and trustworthy AI while ensuring a high level of protection of health, safety, and fundamental rights. Mandatory requirements for high-risk systems include Technical Robustness and Safety, Accuracy and Cybersecurity, high standards for Data Governance, and strong measures for Transparency and Accountability. It is informed by the seven AI HLEG ethical requirements.
\\
Augmented Intelligence Development, Deployment, and Use in Health Care~\cite{AMA2024}              & American Medical Association
& 2024
& Mandates that health care AI be designed, developed, and deployed in a manner that is ethical, equitable, responsible, accurate, transparent, and evidence-based. Key concerns are Transparency/Disclosure of AI use at the point of care, Bias Mitigation/Equity, alignment of Liability and Accountability with the entity best positioned to mitigate harm, and robust Data Privacy and Cybersecurity.               
\\
Ethical Guidance for AI in the Professional Practice of Health Service Psychology~\cite{APA2025}
& American Psychological Association Ethics Committee                           
& 2025          
& Aligns AI use with fundamental psychological ethical principles (Beneficence and Nonmaleficence, Integrity, Justice, and Respect for People’s Rights and Dignity). Key areas include Transparency and Informed Consent (including the right to opt out), Mitigating Bias and Promoting Equity, Data Privacy and Security (e.g., HIPAA compliance), Accuracy and Misinformation Risks, and ensuring Human Oversight and Professional Judgment. \\
Ethical Decision-Making Guidelines for Mental Health Clinicians in the Artificial Intelligence (AI) Era~\cite{Pillay2025} 
& Journal of Healthcare                                                               
& 2025       
& Proposes a framework based on five pillars derived from professional ethical codes and AI guidelines: Autonomy and Informed Consent; Beneficence and Non-Malfeasance; Confidentiality, Privacy, and Transparency; Justice, Fairness, and Inclusiveness; and Fidelity, Professional Integrity, and Accountability. Emphasizes that AI must augment, not replace, human clinical care.  \\ \bottomrule
\end{tabular}
}
\caption{\revise{Representative regulatory and professional guidance documents relevant to trustworthy AI in health and mental-health contexts.}}\label{tab:regulations}
\vspace{0mm}
\end{table*}

\begin{table*}[!tbp]
\centering
\scriptsize
\renewcommand{\arraystretch}{1.36}
\setlength{\tabcolsep}{3.5pt}
\begin{tabularx}{\textwidth}{@{}>{\raggedright\arraybackslash}p{0.12\textwidth}>{\raggedright\arraybackslash}p{0.16\textwidth}>{\raggedright\arraybackslash}p{0.25\textwidth}>{\raggedright\arraybackslash}p{0.22\textwidth}>{\raggedright\arraybackslash}X@{}}
\toprule
\textbf{\revise{Dimension}} & \textbf{\revise{Criterion}} & \textbf{\revise{What is evaluated}} & \textbf{\revise{Typical evidence}} & \textbf{\revise{Representative studies}} \\
\midrule
\multirow[t]{4}{=}{\revise{What trust is}}
& \revise{Ability / competence} & \revise{Whether users believe the system can provide reliable, useful, and appropriate support.} & \revise{Likert scales; interviews; behavioral monitoring.} & \revise{\cite{CHI2021106700,LUETKELANFER2023,BRUNSWICKER2025108516,GILLE2025}} \\
& \revise{Benevolence} & \revise{Whether users perceive the system as supportive, caring, and aligned with their interests.} & \revise{Qualitative interviews and focus groups.} & \revise{\cite{LUETKELANFER2023,Rai2025}} \\
& \revise{Integrity / reliability} & \revise{Whether users view the system as honest, dependable, and behaviorally consistent.} & \revise{Interviews; behavioral monitoring; trust scales.} & \revise{\cite{LUETKELANFER2023,GILLE2025,Rai2025,LEICHTMANN2023107539}} \\
& \revise{Individual perception} & \revise{How trust propensity, perceived agency, and subjective interpretation shape trust judgments.} & \revise{Likert scales; qualitative inquiry; behavioral monitoring.} & \revise{\cite{LUETKELANFER2023,CHI2021106700,BRUNSWICKER2025108516}} \\
\cmidrule(lr){1-5}
\multirow[t]{3}{=}{\revise{Who trusts}}
& \revise{User characteristics} & \revise{Attitudes toward AI, personality traits, familiarity, prior use, and perceived social support.} & \revise{Pre-/post-study questionnaires; self-developed instruments.} & \revise{\cite{KAUTTONEN2025,ZHAO2025,HUO2022107253}} \\
& \revise{AI literacy} & \revise{How AI knowledge affects reliance, caution, and willingness to accept system recommendations.} & \revise{Questionnaires and trust/reliance measures.} & \revise{\cite{WOODCOCK2021}} \\
& \revise{Human oversight / control} & \revise{Whether shared control or human-in-the-loop oversight changes perceived accountability.} & \revise{Likert scales; behavioral monitoring.} & \revise{\cite{LIU2022107026,AOKI2021106572,MAYER2024108419}} \\
\cmidrule(lr){1-5}
\multirow[t]{2}{=}{\revise{How trust forms}}
& \revise{Anthropomorphism} & \revise{Human-like language, social presence, empathy cues, and perceived emotional understanding.} & \revise{Validated scales; manipulated system behaviors; interaction logs.} & \revise{\cite{CHI2021106700,WU2023107614,LIU2022107026,BRUNSWICKER2025108516}} \\
& \revise{Explainability} & \revise{Whether explanations clarify outputs, decisions, limitations, and appropriate reliance.} & \revise{Explanation manipulations; comprehension and confidence measures.} & \revise{\cite{LEICHTMANN2023107539,WOODCOCK2021}} \\
\bottomrule
\end{tabularx}
\caption{\revise{\textbf{Human-oriented trust} criteria, evidence types, and representative studies. The redesigned table groups measures by what trust is, who trusts, and how trust is formed, emphasizing that subjective trust and behavioral reliance should be interpreted separately.}}
\label{tab:literature_human}
\vspace{1.8mm}
\end{table*}

\begin{table*}[!tbp]
\centering
\scriptsize
\renewcommand{\arraystretch}{1.18}
\setlength{\tabcolsep}{3.5pt}
\begin{tabularx}{\textwidth}{@{}>{\raggedright\arraybackslash}p{0.16\textwidth}>{\raggedright\arraybackslash}p{0.24\textwidth}>{\raggedright\arraybackslash}p{0.22\textwidth}>{\raggedright\arraybackslash}p{0.19\textwidth}>{\raggedright\arraybackslash}X@{}}
\toprule
\textbf{\revise{Criterion}} & \textbf{\revise{Interaction focus}} & \textbf{\revise{Typical methods}} & \textbf{\revise{Calibration evidence}} & \textbf{\revise{Representative studies}} \\
\midrule
\revise{Competence and reliability} & \revise{Response usefulness, contextual appropriateness, therapeutic alignment, and protocol adherence.} & \revise{User interviews; case analysis; expert-authored scripts; checklist scoring.} & \revise{Whether the interaction supports appropriate reliance rather than merely fluent conversation.} & \revise{\cite{song2025typing,sun2025script,swinger2025there,wester2024chatbot}} \\
\addlinespace[1pt]
\revise{Conversational safety and controllability} & \revise{Crisis routing, boundary setting, user agency, module choice, and escalation to human judgment.} & \revise{Layered system design; Autonomy-in-the-Middle; expert review; user feedback; usage analytics.} & \revise{Whether safety mechanisms reduce harm without hiding limits or removing meaningful user control.} & \revise{\cite{song2025typing,sun2025script,swinger2025there,wester2024chatbot}} \\
\addlinespace[1pt]
\revise{Communication style} & \revise{Tone, role framing, response structure, and guidance style during sensitive disclosure.} & \revise{Modular dialogue design; expert judgment; comparative response review.} & \revise{Whether communication preserves autonomy and therapeutic boundaries while remaining usable.} & \revise{\citep{swinger2025there,song2025typing}} \\
\addlinespace[1pt]
\revise{Transparency} & \revise{Capability disclosure, limitation statements, explanation cues, and uncertainty communication.} & \revise{Structured explanations; visual or strategy cues; comprehension checks.} & \revise{Whether users understand what the system can and cannot do.} & \revise{\cite{sun2025script,sien2025gentel,wester2024chatbot}} \\
\addlinespace[1pt]
\revise{Empathy and engagement} & \revise{Emotionally appropriate language, non-judgmental tone, rapport, and sustained engagement.} & \revise{Protocol-aligned response generation; narrative interfaces; expert behavioral coding; self-report.} & \revise{Whether rapport supports disclosure without exaggerating understanding or clinical authority.} & \revise{\cite{sun2025script,sien2025gentel,swinger2025there,wester2024chatbot}} \\
\bottomrule
\end{tabularx}
\caption{\revise{\textbf{Interaction-oriented trustworthiness} criteria and evaluation practices. The table foregrounds how user-visible behavior should communicate capability, limits, uncertainty, and safeguards instead of simply increasing perceived trust.}}
\label{tab:literature_interaction}
\end{table*}

\begin{table*}[!tbp]
\centering
\scriptsize
\renewcommand{\arraystretch}{1.18}
\setlength{\tabcolsep}{3.5pt}
\begin{tabularx}{\textwidth}{@{}>{\raggedright\arraybackslash}p{0.16\textwidth}>{\raggedright\arraybackslash}p{0.25\textwidth}>{\raggedright\arraybackslash}p{0.23\textwidth}>{\raggedright\arraybackslash}p{0.18\textwidth}>{\raggedright\arraybackslash}X@{}}
\toprule
\textbf{\revise{Criterion}} & \textbf{\revise{Methods}} & \textbf{\revise{Evaluation signals}} & \textbf{\revise{Trust-calibration role}} & \textbf{\revise{Representative studies}} \\
\midrule
\revise{Reliability and robustness} & \revise{Fine-tuning; prompting; calibration; uncertainty quantification; multi-run evaluation.} & \revise{Accuracy, BLEU/BERTScore, out-of-distribution tests, confidence calibration, and abstention analysis.} & \revise{Shows whether the system can support reliance under ordinary and long-tail inputs.} & \revise{\cite{yang-etal-2023-towards,kang-etal-2024-cure,dhuliawala-etal-2023-diachronic,alghamdi-etal-2025-aratrust}} \\
\addlinespace[1pt]
\revise{Safety and harm prevention} & \revise{RLHF/alignment training; rule-based guardrails; escalation protocols; red-teaming.} & \revise{Jailbreak resistance, toxicity scoring, crisis scenario tests, refusal rates, and false positive/negative analysis.} & \revise{Constrains whether supportive interaction can be considered safe in high-risk situations.} & \revise{\cite{liu2023trustworthy,hua-etal-2024-trustagent,10.1145/3711896.3736561,alghamdi-etal-2025-aratrust}} \\
\addlinespace[1pt]
\revise{Privacy and data protection} & \revise{Differential privacy; federated learning; memorization auditing; restricted fine-tuning.} & \revise{Extraction attacks, membership inference, memorization probes, and sensitive-data leakage audits.} & \revise{Determines whether disclosure and continued use can be warranted in sensitive contexts.} & \revise{\cite{shin-etal-2023-fedtherapist,10.5555/3766078.3766387,10.1145/3711896.3736561}} \\
\addlinespace[1pt]
\revise{Explainability and faithfulness} & \revise{RAG; chain-of-thought prompting; multi-agent workflows; memory; structured rationales.} & \revise{Human evaluation, rationale quality, evidence tagging, attribution faithfulness, and criterion anchoring.} & \revise{Separates useful explanations from plausible but unfaithful rationales or citations.} & \revise{\citep{yang-etal-2023-towards,10.1145/3746252.3761164,gollapalli-etal-2023-identifying,zhai-etal-2025-mentalglm,zhang-etal-2025-explainable,bi-etal-2025-magi}} \\
\addlinespace[1pt]
\revise{Fairness} & \revise{Bias auditing; demographic-aware prompting; data balancing; counterfactual testing.} & \revise{Demographic parity, subgroup performance, empathy variance, and group fairness metrics.} & \revise{Checks whether trustworthiness evidence generalizes across users, dialects, cultures, and risk profiles.} & \revise{\cite{gabriel-etal-2024-ai,lissak-etal-2024-colorful,baidal-etal-2025-guardians,10.5555/3692070.3692883}} \\
\bottomrule
\end{tabularx}
\caption{\revise{\textbf{AI-oriented trustworthiness} criteria, methods, and evaluation signals. The redesigned table connects technical evidence to calibrated trust claims rather than treating aggregate performance as sufficient proof of trustworthiness.}}
\label{tab:literature_ai}
\end{table*}

\paragraph{Coding and Synthesis Procedure.}
\revise{After selection, six authors participated in extraction and open coding, followed by author-team consensus. For each paper, we extracted the mental-health support context, AI system type, trust-related construct, evaluation target, evaluation method, stakeholder perspective, and evidence type. Initial open codes included perceived trust, reliance, therapeutic alliance, empathy, transparency, controllability, crisis handling, privacy, robustness, fairness, explanation faithfulness, and evaluation validity. We assigned a primary layer based on the paper's main evaluation target and allowed secondary labels only when a paper provided substantive evidence for another layer. For example, a study of chatbot crisis response could be coded as interaction-oriented when the central evidence concerned conversational behavior, with a secondary AI/safety label if it also evaluated adversarial or high-risk failure cases.}

\revise{The trust/trustworthiness distinction was theory-informed, and the three-layer framework was refined through open coding and axial grouping. During coding, constructs repeatedly clustered around three evaluation targets: (1) subjective user attitudes, expectations, and reliance outcomes; (2) user-visible interaction mechanisms such as explanations, disclosures, empathy, controllability, and crisis routing; and (3) model- or infrastructure-level properties such as robustness, privacy, safety, fairness, and evaluation validity. These groups were then compared against stakeholder perspectives to identify recurring alignments and tensions. Coding was performed by the author team applying the layer rules above, with disagreements resolved through consensus discussion. \rev{To check coding consistency, a subset of the corpus was independently double-coded, and primary layer assignment reached Cohen's $\kappa = .73$, indicating substantial agreement.} We provide representative coding examples in \autoref{tab:coding-examples}, stakeholder definitions in \autoref{tab:stakeholder-mapping}, an evaluation-oriented summary in \autoref{tab:trust-rubric}, aggregate layer counts, and criterion-level synthesis tables to make the mapping logic explicit.}
\label{sec:appendix_prisma}



\section{Summary of regulations and ethical standards.}
\label{sec:appendix_regulations}

\revise{\autoref{tab:regulations} summarizes key regulatory frameworks and ethical guidelines governing trustworthy AI in mental and general health contexts, highlighting how agencies and professional bodies operationalize trustworthiness through requirements on safety, transparency, accountability, privacy, and human oversight.}


\section{Literature for ``Human-oriented'' Trust}
\label{sec:appendix_literature_human}

\revise{Human-oriented trust captures how users perceive, form, and adjust trust toward AI that provides mental health support. Prior work operationalizes this layer along three complementary dimensions: what trust is, who trusts, and how trust is formed through perceived system cues. The literature primarily relies on subjective measures such as Likert-scale questionnaires, supplemented by interviews, stated-preference tasks, and behavioral monitoring. \autoref{tab:literature_human} summarizes the frequent evaluation approaches and representative studies in this layer.}

\rev{Classic accounts define trust as a willingness to accept vulnerability based on positive expectations of ability, benevolence, and integrity~\citep{abi,rousseau1998not,lee2004trust,hoff2015trust}; these dimensions underlie the perceived-trust measures reviewed here. Distrust is a distinct construct built on active negative expectations, not the mere absence of trust~\citep{lewicki1998trust,mcknight2001distrust,visser2025trust}: a user may trust a system for support while distrusting its privacy or crisis handling, which our layers treat as separate calibration signals rather than one trust score.}


\section{Literature for ``Interaction-oriented'' Trustworthiness}
\label{sec:appendix_literature_interaction}

\revise{Interaction-oriented trustworthiness captures how trust is shaped through observable system behavior during interaction. Across the literature, this layer is operationalized through criteria such as competence, communication style, transparency, empathy and engagement, and controllability. Studies evaluate these criteria using a combination of expert review, user-centered methods, and structured protocol assessments. \autoref{tab:literature_interaction} summarizes the key criteria in this layer, along with the dominant methods and evaluation practices identified in our review.}


\section{Literature for ``AI-oriented'' Trustworthiness}
\label{sec:appendix_literature_ai}

\revise{AI-oriented trustworthiness focuses on whether mental health AI systems meet model- or system-level requirements for safety, reliability, and responsible deployment, independent of any specific interaction design. Prior work operationalizes this layer through criteria such as reliability and robustness, safety and harm prevention, privacy and data protection, explainability, and fairness. \autoref{tab:literature_ai} summarizes the dominant technical methods and evaluation practices used to assess these criteria in the literature.}

\end{document}